\documentclass[conference]{IEEEtran}
\IEEEoverridecommandlockouts
\usepackage[ruled,linesnumbered,noend]{algorithm2e}

\usepackage{cite}
\usepackage{paralist}
\usepackage{amsmath,amssymb,amsfonts}
\usepackage{graphicx}
\usepackage{textcomp}
\usepackage{xcolor}
\usepackage{xspace}
\usepackage{comment}
\usepackage{algorithmic}
\usepackage{subcaption} 
\usepackage{pdfpages}
\usepackage{url}
\usepackage{outlines}
\usepackage{amsthm}
\usepackage{amsmath}
\usepackage{marvosym}
\usepackage{wasysym}
\usepackage{verbatim}
\usepackage{float}
\usepackage{hyperref}
\usepackage{multirow}
\usepackage{cite}
\usepackage[capitalize]{cleveref}
\usepackage[per-mode=symbol,detect-all]{siunitx}
\usepackage{graphics}
\usepackage{amssymb}
\usepackage{enumitem}

\usepackage{booktabs}
\usepackage{colortbl}
\usepackage{ifthen}
\usepackage{mathdots}
\usepackage{arydshln}
\usepackage[outline]{contour}

\usepackage{wrapfig}
\usepackage{MnSymbol}
\usepackage{placeins}
\usepackage{diagbox}
\usepackage{listings}
\usepackage{pdfpages}

\usepackage{xcolor,pifont}
\newcommand*\colourcheck[1]{%
	\expandafter\newcommand\csname #1check\endcsname{\textcolor{#1}{\ding{52}}}%
}
\newcommand*\colourxmark[1]{%
	\expandafter\newcommand\csname #1xmark\endcsname{\textcolor{#1}{\ding{56}}}%
}
\colourcheck{green}
\colourxmark{red}

\definecolor{navy}{RGB}{0,0,128}

\newcommand{\X}{\mathcal{X}}

\usepackage{tikz,ifthen,pgfplots}
\usetikzlibrary{arrows,trees,backgrounds,automata,shapes,decorations,plotmarks,fit,calc,positioning,shadows,chains}
\usetikzlibrary{quotes,arrows.meta}
\tikzstyle{every pin edge}=[<-,shorten <=1pt]
\tikzstyle{neuron}=[circle,fill=black!25,minimum size=17pt,inner sep=0pt]
\tikzstyle{input neuron}=[neuron, fill=green!50]
\tikzstyle{output neuron}=[neuron, fill=red!50]
\tikzstyle{hidden neuron}=[neuron, fill=blue!50]
\tikzstyle{small neuron}        =[hidden neuron, draw, minimum size=15pt]
\tikzstyle{small input neuron}  =[input neuron , draw, minimum size=15pt]
\tikzstyle{small output neuron} =[output neuron, draw, minimum size=15pt]
\tikzstyle{annot} = [text width=4em, text centered]
\tikzstyle{nnedge} = [-{stealth},shorten >=0.1cm, shorten <=0.05cm,line width=0.8pt,black]
\tikzstyle{edge} = [->,line width = 0.3pt, shorten >=0.2cm]
\tikzstyle{edgeWide} = [->,line width = 2pt, , shorten >=0.2cm]
\usetikzlibrary{calc}

\tikzset{every picture/.style={line width=0.75pt}} 
\tikzstyle{BadSquare}=[rectangle,fill=red!30!white,minimum size=25pt,inner 
sep=0pt]
\tikzstyle{InitSquare}=[rectangle,fill=green!30!white,minimum size=25pt,inner 
sep=0pt]

\newcommand{\sat}{\texttt{SAT}\xspace}

\newcommand{\unsat}{\texttt{UNSAT}\xspace}

\usepackage{amsthm}
\newtheorem{lemma}{Lemma}

\usepackage{mdframed}

\newmdtheoremenv{definition}{Definition}

\usepackage{bussproofs}
\usepackage{gensymb}

\newif\ifcomments
\commentstrue

\newif\ifoutline
\outlinefalse

\newif\iflong
\longtrue

\renewcommand{\paragraph}[1]{\vspace{1mm}\noindent{\bf #1}\ }

\newcommand{\states}{\mathcal{X}\xspace}
\newcommand{\state}{x\xspace}
\newcommand{\traj}{\tau\xspace}
\newcommand{\control}{u\xspace}
\newcommand{\controls}{\mathcal{U}\xspace}
\newcommand{\trans}{f\xspace}
\newcommand{\certfun}{V\xspace}
\newcommand{\ctrlfun}{\pi\xspace}

\IEEEoverridecommandlockouts
\DeclareRobustCommand{\IEEEauthorrefmark}[1]{\smash{\textsuperscript{\footnotesize
 #1}}}

	\author{
		\IEEEauthorblockN{
			Udayan Mandal\IEEEauthorrefmark{1},
			Guy Amir\IEEEauthorrefmark{2},
			 Haoze Wu\IEEEauthorrefmark{1},
			 Ieva Daukantas\IEEEauthorrefmark{3},
            Fletcher Lee Newell\IEEEauthorrefmark{1},
			Umberto J. Ravaioli\IEEEauthorrefmark{4},\\
			 Baoluo Meng\IEEEauthorrefmark{5},
			 Michael Durling\IEEEauthorrefmark{5},
          Milan Ganai\IEEEauthorrefmark{1},
          Tobey 
          Shim\IEEEauthorrefmark{1},
          Guy Katz\IEEEauthorrefmark{2},
			and Clark Barrett\IEEEauthorrefmark{1}
		}
  
		\IEEEauthorblockA{			\IEEEauthorrefmark{1}Stanford University, 
			\IEEEauthorrefmark{2}The Hebrew University of Jerusalem, 
   \IEEEauthorrefmark{3}IT University of Copenhagen,
\IEEEauthorrefmark{4}Google, 
   \IEEEauthorrefmark{5}GE Aerospace Research
		}
	}

\def\BibTeX{{\rm B\kern-.05em{\sc i\kern-.025em b}\kern-.08em
    T\kern-.1667em\lower.7ex\hbox{E}\kern-.125emX}}

\title{Safe and Reliable Training\\ of Learning-Based Aerospace Controllers \\
{}
}

\author{
\IEEEauthorblockN{Udayan Mandal}
\IEEEauthorblockA{\textit{Center for AI Safety} \\
\textit{Stanford University}\\
Stanford, USA \\
udayanm@stanford.edu}
\and
\IEEEauthorblockN{Guy Amir}
\IEEEauthorblockA{\textit{School of CS \& Engineering} \\
\textit{The Hebrew University of Jerusalem}\\
Jerusalem, Israel \\
guyam@cs.huji.ac.il}
\and
\IEEEauthorblockN{Haoze Wu}
\IEEEauthorblockA{\textit{Center for AI Safety} \\
\textit{Stanford University}\\
Stanford, USA \\
haozewu@stanford.edu}
\and
\IEEEauthorblockN{Ieva Daukantas}
\IEEEauthorblockA{\textit{Department of Computer Science} \\
\textit{IT University of Copenhagen}\\
Copenhagen, Denmark \\
daukantas@itu.dk}
\and
\IEEEauthorblockN{Fletcher Lee Newell}
\IEEEauthorblockA{\textit{Center for AI Safety} \\
\textit{Stanford University}\\
Stanford, USA \\
flnewell@stanford.edu}
\and
\IEEEauthorblockN{Umberto Ravaioli}
\IEEEauthorblockA{
\textit{Google}\\
Mountain View, USA \\
uravaioli@google.com}
\and
\IEEEauthorblockN{Baoluo Meng}
\IEEEauthorblockA{\textit{GE Aerospace Research} \\
Niskayuna, USA \\
baoluo.meng@ge.com}
\and
\IEEEauthorblockN{Michael Durling}
\IEEEauthorblockA{\textit{GE Aerospace Research} \\
Niskayuna, USA \\
durling@ge.com}
\and
\IEEEauthorblockN{Kerianne Hobbs}
\IEEEauthorblockA{\textit{Air Force Research Laboratory} \\
\textit{US Air Force}\\
Dayton, USA \\
kerianne.hobbs@afrl.af.mil}
\and
\IEEEauthorblockN{Milan Ganai}
\IEEEauthorblockA{\textit{Department of Computer Science} \\
\textit{Stanford University}\\
Stanford, USA \\
mganai@stanford.edu}
\and
\IEEEauthorblockN{Tobey Shim}
\IEEEauthorblockA{\textit{Department of Data Science} \\
\textit{Stanford University}\\
Stanford, USA \\
tshim24@stanford.edu}
\and
\IEEEauthorblockN{Guy Katz}
\IEEEauthorblockA{\textit{School of CS \& Engineering} \\
\textit{The Hebrew University of Jerusalem}\\
Jerusalem, Israel \\
guykatz@cs.huji.ac.il}
\and
\IEEEauthorblockN{Clark Barrett}
\IEEEauthorblockA{\textit{Center for AI Safety} \\
\textit{Stanford University}\\
Stanford, USA \\
barrett@stanford.edu}
}

\begin{document}

\maketitle

\begin{abstract}
In recent years, deep reinforcement learning (DRL) approaches have generated highly successful controllers for a myriad of complex domains. However, the opaque nature of these models limits their applicability in aerospace systems and sasfety-critical domains, in which a single mistake can have dire consequences.
In this paper, we present novel advancements in both the training and verification of DRL controllers, which can help ensure their safe behavior. 
We showcase a design-for-verification approach utilizing $k$-induction and demonstrate its use in verifying liveness properties. 
In addition, we also give a brief overview of neural Lyapunov Barrier certificates and summarize their capabilities on a case study. Finally, we describe several other novel reachability-based approaches which, despite failing to provide guarantees of interest, could be effective for verification of other DRL systems, and could be of further interest to the community.
\end{abstract}


\section{Introduction}
\label{sec:Introduction}

Deep reinforcement learning (DRL) has gained significant popularity in recent years, reaching state-of-the-art performance in various domains. One such domain is aerospace systems, in which DRL models are under consideration for replacing years-old software by learning to efficiently control airborne platforms and spacecraft. However, although they perform well empirically, DRL systems have an opaque decision-making process, making them challenging to reason about. More importantly, this opacity raises critical questions about safety and security (e.g., \emph{How can we ensure that the spacecraft will never violate a velocity constraint? Will it always reach its destination?}) which are difficult to answer. These reliability concerns are a significant obstacle to deploying DRL controllers in real-world systems, where even a single mistake cannot be tolerated. 

To cope with this urgent need, a myriad of DRL \emph{training techniques} have been put forth in recent years to enhance the performance of such systems. However, these current approaches suffer from two main drawbacks: (i) they are usually not geared towards improving safety and reliability (which is key in aerospace systems); and (ii) they are heuristic in nature and do not afford any formal guarantees. At the same time, the formal methods community has been developing methods for formally and rigorously assessing the reliability of DRL systems. However, although such methods are useful for identifying whether a system is safe, they are typically not incorporated into the DRL training process, but are rather used only afterwards.

In this work, we begin bridging this gap by proposing a novel design-for-verification approach that can be incorporated during the DRL training process.  Our approach both modifies the training loop to be more verification-friendly and also utilizes formal verification (in our case, $k$-induction), to ensure the correctness of the training.
We also report a summary of our recent efforts to use Neural Lyapunov Barrier certificates~\cite{dawson2023safe} to generate DRL agents that not only perform well on large batches of data, but also meet rigorous correctness criteria as measured by state-of-the-art verification tools. 

Finally, we introduce additional novel reachability-based approaches for providing safety and liveness guarantees about a DRL system. These approaches are derived from prior work on backward-tube reachability, forward-tube reachability, and abstraction-based reachability methods. Moreover, these approaches all follow a similar paradigm: the reachable space covered by all possible paths from the starting state space is over-approximated using a verification engine, and safety and liveness properties are checked over this over-approximated state space. 
To demonstrate the usefulness of our approaches, we apply them to a benchmark satellite-control model developed in collaboration with industry partners (GE Aerospace Research and the U.S. Air Force). We demonstrate that liveness can be verified using our $k$-induction approach. Additionally, as a point of comparison, we showcase that the certificate-based approach is indeed able to generate a controller that provably behaves safely. Notably, the problem setting and controller complexity are beyond that acheived in previous work on formally verified controllers. 


The other reachability-based methods fail on this benchmark.  However, we believe that these failed attempts: (i) demonstrate the merits of our successful approaches in handling complex, nontrivial properties; (ii) can be of value to the community, by shedding light on vulnerabilities of alternate methods; and (iii) could be potentially successful when applied over different DRL systems. 

We view this work as an important step towards the safe and reliable deployment of DRL controllers in real-world systems, especially in the complex domain of avionics. We additionally hope that our work will further motivate additional research in neural network verification, DRL safety, and specifically, their role in the important domain of DRL-controlled aerospace systems. 

The rest of the paper is organized as follows. In Sec.~\ref{sec:background}, we cover background on deep learning, DRL, and verification, and we also introduce Neural Lyapunov Barrier functions.
In Sec.~\ref{sec:dockingproblem}, we introduce our benchmark problem, a 2D spacecraft docking challenge.
We subsequently introduce our k-induction technique in Sec.~\ref{sec:kinduction}, and we present alternative verification approaches in Sec.~\ref{sec:certificate}. \footnote{Code for these approaches is available at: \href{https://github.com/NeuralNetworkVerification/artifact-dasc-docking}{github.com/NeuralNetworkVerification/artifact-dasc-docking}}
Finally, we conclude in Sec.~\ref{sec:Conclusion}.



\section{Preliminaries and Related Work}
\label{sec:background}

\subsection{Safety and Liveness}

In this paper, we are interested in obtaining DRL controllers that satisfy safety and liveness properties~\cite{AlSc87} in discrete-time settings.

%
%

\paragraph{Safety.}
In a sequence satisfying a safety property, \emph{a bad state is never reached}. For the set of system states $\states$, let $\traj \subseteq \states^*$ be the set of potential system trajectories.  We say a trajectory $\alpha$ satisfies \emph{safety} property $P_{1}$ if and only if each state in $\alpha$ satisfies property $P_{1}$. More formally:
\begin{equation}
\forall\, \alpha:\alpha \in \traj. \forall\,\state  \in \alpha.\: \state\vDash P_{1}.
\end{equation}
Finite-length trajectories terminating in a ``bad'' state (where $P_{1}$ does not hold) constitute the set of trajectories in violation of the safety property.

\paragraph{Liveness.}
On the other hand, a liveness property indicates \emph{a good state is eventually reached}. 
A \emph{liveness} property $P_{2}$ is satisfied by trajectory $\alpha$ if and only if there exists a state $\state$ in $\alpha$ where $P_{2}$ holds.  Defining $\traj^{\infty}$ as the set of infinite-length trajectories, we formally specify liveness property $P_{2}$ as:
\begin{equation}
\forall\, \alpha :\alpha \in \tau^{\infty}.\:\exists\, \state \in \alpha.\: \state\vDash P_{2}.
\end{equation}
Infinite-length trajectories which contain no ``good" states (i.e., no states where $P_{2}$ holds) constitute the set of trajectories in violation of the liveness property.

\subsection{DNNs, DNN Verification, and Dynamical Systems.} 

\paragraph{Deep Learning.} 
Deep neural networks (DNNs) consist of layers of neurons that perform some (usually nonlinear) transformation of the input~\cite{GoBeCo16}. 
In this paper, we investigate deep reinforcement learning (DRL), where we train a DNN to obtain a \emph{policy}, which maps states to actions that control a system~\cite{Li17}.

\paragraph{DNN Verification.}
Given \begin{inparaenum}[(i)]
		\item a trained DNN (e.g., a DRL agent) $N$;
		\item a pre-condition $P$ on the DNN's inputs, limiting the input assignments; and	
		\item a post-condition $Q$ on the DNN's outputs%
\end{inparaenum},
 the goal of DNN verification is to determine whether the property $P(\state)\rightarrow Q(N(\state))$ holds for any neural network input $\state$. In many DNN verifiers (a.k.a., \emph{verification engines}), this task is equivalently reduced to determining the satisfiability of the formula $P(\state)\land \neg Q(N(\state))$. If the formula is satisfiable (\sat), then there is an input that satisfies the pre-condition and violates the post-condition, which means the property is violated. On the other hand, if the formula is unsatisfiable (\unsat), then the property holds. It has been shown~\cite{KaBaDiJuKo17} that verification of piece-wise-linear DNNs is NP-complete.
 In recent years, the formal methods community has put forth 
	various techniques for verifying and improving DNN reliability~\cite{
		 AmMaZeKaSc23,AmZeKaSc22,
		UsGoSuNoPa21, AmCoYeMaHaFaKa23,BaAmCoReKa23,AlAvHeLu20,CaKoDaKoKaAmRe22,CoAmKaFa24}. 
	These techniques include SMT-based methods~\cite{HuKwWaWu17, AmWuBaKa21,
		KaHuIbJuLaLiShThWuZeDiKoBa19, KuKaGoJuBaKo18}, optimization-based methods~\cite{LoMa17,Eh17,TjXiTe19,
		BuTuToKoMu18}, methods based on abstraction-refinement~\cite{ElGoKa20, PrAf20, AnPaDiCh19, 
		SiGePuVe19, OsBaKa22,ElCoKa22,CoElBaKa22}, methods based on shielding~\cite{KoLoJaBl20,RoAmCoSaKa24, CoAmRoSaKaFo24}, and more.

\paragraph{Discrete-Time Dynamical Systems.}
We consider discrete-time dynamical systems, particularly systems whose trajectories satisfy the equation:
\begin{equation}
    \state_{t+1} = \trans(\state_t, \control_t),
\end{equation}
in which the \emph{transition function} $\trans$ takes as inputs the current state $\state_t\in\states$ and a control $\control_t\in\controls$ and produces as output the subsequent state $\state_{t+1}$. To control these systems, we employ a policy $\ctrlfun:\states \rightarrow \controls$ that takes in a state $\state\in\states$ and outputs a control action $\control = \ctrlfun(\state)$. 
In DRL, the controller $\ctrlfun$ is realized by a trained DNN agent. These learning-based controllers have proven to be effective in many real-world settings including robotics~\cite{dawson2023safe}, biomedical systems~\cite{de2022intelligent}, and energy management~\cite{huang2021neural}, due to their expressive power and ability to generalize to unseen, complex environments~\cite{TalSobKir2019}.



\subsection{Control Lyapunov Barrier Functions}

The problem of verifying safety and liveness properties in a dynamical system can be solved by finding a function $\certfun: \states\mapsto \mathbb{R}$ with certain properties. 
Control theory identifies two fundamental types of functions~\cite{li2023survey}.

\paragraph{Lyapunov Functions.}
Lyapunov functions, a.k.a., \emph{Control Lyapunov functions}, capture the energy level at a particular state: 
over time, energy is dissipated along a trajectory until the system attains zero-energy equilibrium~\cite{lyapunovbook}. 
Lyapunov functions can guarantee 
\textit{asymptotic stability}, which ensures the system eventually converges to some goal state (thereby satisfying a liveness property). 
Lyapunov functions must be ($i$) equal to $0$ at equilibrium, ($ii$) strictly positive at all other states; and ($iii$) monotonically  decreasing~\cite{NLC, Chang2021,ganai2023learning}.

\paragraph{Barrier Functions.}
Barrier functions~\cite{cbfsurvey}, a.k.a., \emph{Control Barrier Functions}, 
 guarantee that a system never enters an unsafe region (i.e., a ``bad" state) in the state space.  This is achieved by setting the function value to be above some threshold for unsafe states and then verifying that the system can never transition to a state where the function is above the threshold~\cite{cbfqptac, ames_robust, basile1969controlled}. Previous work~\cite{qin2022quantifying, qin2021learning, tong2023enforcing, yu2023sequential} demonstrates how to obtain Barrier functions for various safety-critical tasks such as pedestrian avoidance, neural radiance field-based obstacle navigation~\cite{mildenhall2021nerf}, and multi-agent control.

\paragraph{Control Lyapunov Barrier Functions.}
Often, it is necessary to ensure both safety and liveness properties simultaneously. 
In such cases,  we can employ a \emph{Control Lyapunov Barrier Function} (CLBF), which integrates the properties and guarantees of both Control Lyapunov functions and Control Barrier functions~\cite{dawson2022safe}. 
CLBFs can solve reach-while-avoid tasks~\cite{EdwPerAba2023}, which we discuss next. 

%
%

\paragraph{Reach-while-Avoid Tasks.}
The goal of \emph{Reach-while-Avoid} (RWA) tasks is to find a controller $\pi$ for a dynamical system such that every trajectory $\{\state_1, \state_2 ... \}$ produced under this controller (i) never enters an unsafe (``bad'') state; and (ii) eventually enters a goal (``good") region or state. 
%
%
%
We can formally define the problem as:

\begin{definition}[Reach-while-Avoid Task]
\label{def:reachavoid}
    \textcolor{white}{ }\\
   \textbf{Input: } 
   A dynamical system with a set of initial states $\states_I\subseteq\states$, a set of goal states $\states_G\subseteq \states$, and a set of unsafe states $\states_U\subseteq \states$, where $\states_I\cap \states_U =\emptyset$ and $\states_G \cap \states_U = \emptyset$\\
   \textbf{Output: }
   A controller $\pi$ such that for every trajectory $\traj=\{\state_1, \state_2 ... \}$ satisfying $\state_1 \in \states_I$:
    \begin{enumerate}
        \item \textbf{Reach:} $\exists\, t \in \mathbb{N}.\: x_t \in \states_G$
        \item \textbf{Avoid:} $\forall\, t\in \mathbb{N}.\: x_t\not\in \states_U$ 
    \end{enumerate}
\end{definition}

Some solutions for RWA tasks rely on control theoretic principles. The approach in~\cite{dawson2022safe} trains Lyapunov and Barrier certificates to solve RWA tasks. Hamilton-Jacobi (HJ) reachability-based methods~\cite{hjreachabilityoverview}) have also been employed to solve RWA tasks~\cite{fisac2015reach,hsu2021safety,so2023solving}. Safe DRL is closely connected to RWA, with its goal being to maximize cumulative rewards while minimizing costs along a trajectory~\cite{brunke2022safe}. It has been solved with both Lyapunov/Barrier methods~\cite{yang2023model, Chow2018} and HJ reachability methods~\cite{yu2022reachability, ganai2023iterative}. 

\subsection{Other Verification Approaches}
\paragraph{Reachability Analysis.}
Reachability analysis methods aim to define and compute the set of final reachable states and then verify that this set (i) does not include any bad states, and (ii) is contained within the goal region. Reachability methods include forward-tube and backward-tube verification~\cite{gupta2020safety}, which either propagate states forward from the starting set or backward from the goal set. Other related work in reachability analysis includes hybrid system verifiers~\cite{ivanov2018verisig}, growing the set of reachable states over a discrete action space~\cite{julian2019reachability}, approximating reachable states during forward and backward reachability~\cite{742898}, and reformulating the dynamics of a system for easier reachability verification~\cite{gates2023scalable}.

\paragraph{Bounded Model Checking and $k$-induction.}
Bounded model checking uses a symbolic analysis over $k$ copies of a system to check whether a bad state is reachable in $k$ or fewer steps from the starting set of states.  $k$-induction is similar, except that it starts from an arbitrary state and can thus be used to prove that a bad state is never reached.
Bounded model checking has been explored in the WhiRL tool~\cite{Eliyahu2021VerifyingLS} using the neural network verifier Marabou~\cite{KaHuIbJuLaLiShThWuZeDiKoBa19,wu2024marabou}. \cite{9046094} implements another tool for checking adversarial cases and coverage using bounded model checking for artificial neural networks. WhiRL 2.0~\cite{AmScKa21} adds $k$-induction capabilities to WhiRL.

\paragraph{Design-for-Verification.} Design-for-verification broadly encompasses any method which aims to modify the design and training process to make verification easier. The Trainify framework~\cite{10.1007/978-3-031-13185-1_10} uses a CEGAR-based approach to grow an easily verifiable state space by repeatedly retraining the DRL system.  \cite{9287929} motivates an optimized DRL training approach to reduce the number of safety violations, easing formal verification. This approach was also implemented in Marabou~\cite{KaHuIbJuLaLiShThWuZeDiKoBa19,wu2024marabou}.




\section{2D Docking Problem}
\label{sec:dockingproblem}

We adopt as a motivating case study benchmark the 2D docking problem presented in~\cite{ravaioli2022safe}.  The goal is to train a DRL controller to safely navigate a deputy spacecraft to a chief spacecraft within two-dimensional space. The reference frame is defined such that the chief spacecraft is always at the origin $(0,0)$.  The state of the deputy spacecraft is $\boldsymbol{\state}=[x, y, \dot{x}, \dot{y}]$, where $(x,y)$ are the position of the spacecraft and $(\dot{x}, \dot{y})$ are the respective directional velocities.

\subsection{Dynamics}
The system dynamics are defined according to the linearly-approximate Clohessy-Wiltshire relative orbital motion equations in a non-inertial Hill’s reference frame~\cite{clohessy1960terminal, hill1878researches}. The control input to the system is $\boldsymbol{\control}=[F_x, F_y]$, where $F_x$ and $F_y$ are the thrust forces applied to the deputy spacecraft in the $x$ and $y$ directions. We follow~\cite{ravaioli2022safe}, setting the deputy spacecraft mass to $m=12$ kg and the mean motion to $n=0.001027$ rad/s. The continuous time state dynamics of the system are given by the following differential equations:

\begin{align}
\dot{\boldsymbol{\state}} &= [\dot{x}, \dot{y}, \ddot{x}, \ddot{y}] \\
\ddot{x} &=  2n\dot{y} + 3n^2x + \frac{F_x}{m}\\
\ddot{y} &= -2n\dot{x} + \frac{F_y}{m}
\end{align}

Integration using a discrete time step $T$ yields a closed-form next-state function.
Given a state $\boldsymbol{\state} = [x, y, \dot{x}, \dot{y}]$ and control inputs $\boldsymbol{u} = [F_x, F_y]$, the spacecraft's next state $\boldsymbol{\state}_i' = [x', y', \dot{x}', \dot{y}']$ after an elapsed time $T$ is:

\begin{align}
\begin{split}\label{eq:cw1}
x' &= (\frac{2 \dot{y}}{n} + 4 x + \frac{F_x}{m n^2}) + (\frac{2 F_y}{mn})
+ (- \frac{F_x}{m n^2} - \frac{2 \dot{y}}{n} - 3 x) \\
&\qquad \cdot \cos{(nT)} + (\frac{-2 F_y}{m n^2} + \frac{\dot{x}}{n})\sin{(nT)}
\end{split}
\\[2ex]
\begin{split}
y' &= (-\frac{2\dot{x}}{n} + y + \frac{4 F_y}{m n^2}) + (\frac{-2 F_x}{mn} - 3\dot{y} - 6nx)T + -\frac{3F_y}{2m}t^2 \\
&\qquad+ (-\frac{4F_y}{mn^2} + \frac{2\dot{x}}{n})\cos{(nT)} + (\frac{2F_x}{mn^2} + \frac{4\dot{y}}{n} + 6x)\\
&\qquad \cdot \sin{(nT)}
\end{split}
\end{align}
\begin{align}
\begin{split}
\dot{x}' &= (\frac{2F_x}{mn}) + (\frac{-2F_y}{mn} + x)\cos{(nT)} + (\frac{F_x}{mn} + 2\dot{y} \\
&\qquad + 3nx) \sin{(nT)}\\
\end{split}
\\[2ex]
\begin{split}\label{eq:cw4}
\dot{y}' &= (\frac{-2F_x}{mn} - 3\dot{y} - 6nx) + (-\frac{3 F_y}{m})T + (\frac{2F_x}{mn} + 4\dot{y} \\
&\qquad + 6nx)\cos{(nT)} + (\frac{4 F_y}{mn} - 2 \dot{x})\sin{(nT)}
\end{split}
\end{align}
\subsection{Liveness —-- Docking Region}
\label{sec:liveness}
The problem as given in~\cite{ravaioli2022safe} defines a \emph{docking region} which is a circle of radius $0.5$ meters centered at the origin.  The goal is for the deputy spacecraft to eventually enter this region.  To simplify the verification query, it is easier to use linear bounds for the goal region, so we use a square centered at the origin with sides parallel to the axes of length $0.7$ meters (note that this square fits inside the docking region of~\cite{ravaioli2022safe}). 
 Formally, our liveness condition is:
\begin{equation}
    \forall \alpha : \alpha\in\tau^{\infty}.\: \exists t.\: |\alpha_{t}.x| \leq 0.35\land |\alpha_{t}.y| \leq 0.35,
\end{equation}
\noindent
where $\alpha_t$ is the state at time $t$ in trajectory $\alpha$, and $\alpha_t.x$ and $\alpha_t.y$ are the $x$ and $y$ components of $\alpha_t$.

\subsection{Safety --- Velocity Threshold}
\label{sec:safety}
To minimize the risk to both spacecraft, a safety constraint is imposed on the magnitude of the velocity of the deputy spacecraft.  The constraint depends on the distance from the deputy.  Formally, ~\cite{ravaioli2022safe} requires the following state invariant:
\begin{equation}
    \sqrt{\dot{x}^2 + \dot{y}^2} \leq 0.2 + 2n \sqrt{x^2+y^2}
    \label{eq:safe}
\end{equation}
We therefore define the unsafe region to be the negation of \eqref{eq:safe}.

\newcommand{\directions}{\ensuremath\mathit{directions}\xspace}

Again, we desire to instead use a linear constraint in order to be compatible with our formal tools.  We use the Euclidean norm approximation of~\cite{camino2019linearization}, which approximates the norm by projecting it onto vectors in all different directions and taking the one with the maximum magnitude.  We use the two inequalities:
\begin{align}
\begin{split}
    \max_{i\in [1,n_{\directions}]} (u_1 \cdot \cos(\frac{2(i-1)\pi}{n_{\directions}}) + u_2 \\
    \cdot \sin(\frac{2(i-1)\pi}{n_{\directions}})) \leq \sqrt{u_1^2+u_2^2}
\end{split} 
\end{align}
and
\begin{align}
\begin{split}
    \frac{1}{cos(\pi/n_{\directions})} \max_{i\in [1,n_{\directions}]} (u_1 \cdot \cos(\frac{2(i-1)\pi}{n_{\directions}}) \\
    +u_2 \cdot \sin(\frac{2(i-1)\pi}{n_{\directions}})) \geq \sqrt{u_1^2+u_2^2},
\end{split}
\end{align}
where $n_{\directions}$ is a positive integer.  Larger values of $n_{\directions}$ yield more precise approximations.
We can simplify this by noting that:
\[ \sqrt{u_1^2+u_2^2} = \sqrt{|u_1|^2+|u_2|^2},\]
and then focusing our search only on vectors in the first quadrant.  Assuming $n_{\directions}$ is a multiple of 4, we get:
\begin{align}
\label{eq:underapprox}
\begin{split}
\text{under}(u_1, u_2)&=
    \max_{i\in [1,n_{\directions}/4+1]} (|u_1| \cdot \cos(\frac{2(i-1)\pi}{n_{\directions}})\\
    &\qquad + |u_2| \cdot \sin(\frac{2(i-1)\pi}{n_{\directions}}))\\
    &\leq \sqrt{u_1^2+u_2^2}
\end{split}
\end{align}
and 
\begin{align}
\label{eq:overapprox}
\begin{split}
    \text{over}(u_1, u_2)&=
    \frac{1}{cos(\pi/n_{\directions})} \max_{i\in [1,n_{\directions}/4+1]} (|u_1| \\
    &\qquad \cdot \cos(\frac{2(i-1)\pi}{n_{\directions}}) + |u_2| \cdot \sin(\frac{2(i-1)\pi}{n_{\directions}}))\\ 
    &\geq \sqrt{u_1^2+u_2^2}.
\end{split}
\end{align}
Using these constraints, we can over-approximate the unsafe region as
\begin{align}
\label{eq:over_unsafe}
    \text{over}(\dot{x}_t, \dot{y}_t) > 0.2 + 2n \cdot \text{under}(x_t, y_t).
\end{align}
\noindent
This is a piece-wise linear constraint.  Moreover, both the absolute value function and the maximum function can be easily encoded in neural network verification tools such as Marabou.  In our experiments, we use $n_{\directions}=400$.

\subsection{DNN Setup}\label{subsec:DNN_setup}
As in~\cite{ravaioli2022safe}, we use Ray RLib's Proximal Policy Optimization (PPO) reinforcement learning algorithm to learn the system dynamics, but we make four important alterations to improve downstream verification, part of our \emph{design for verification} scheme.
\subsubsection{Scenario Regions} To improve performance near the docking region, we reduce the docking distance during training from 0.5 meters to 0.25 meters.  We also simplify the problem by reducing the initial position of the deputy spacecraft from a radius of 150 meters to only 5 meters.  Scaling back up to larger initial positions is part of an ongoing research effort.

\subsubsection{Speed Observations} We limit the observations of the agent to its $x$ and $y$ positions and respective $\dot{x}$ and $\dot{y}$ velocities, eliminating the agent's observations of its current speed and the distance-dependent velocity constraint described in Equation \ref{eq:safe}.  This makes it less likely that irregular trajectories will be learned because of observations of the safety constraint.  As a result, liveness verification becomes easier.
\subsubsection{Distance Reward} We keep the rewards relating to success or failure, the safety constraint, and delta-v as presented in~\cite{ravaioli2022safe}, but we alter the distance change reward to use the $L^1$ norm of the position of the deputy --- i.e., the Manhattan distance from the deputy to the chief, rather than the nonlinear $L^2$ norm. This is to match the induction invariant described in Section~\ref{sec:kinduction}.  To account for the new distance metric and previously-described smaller initial distances, we developed a novel reward function for distance change:
\begin{align}
    R^{d_{new}}_t &= 2 \left( e^{-a_1d^m_{t}} - e^{-a_1d^m_{t-1}} \right) + 2 \left( e^{-a_2d^m_{t}} - e^{-a_2d^m_{t-1}} \right),
\end{align}
where $d^m_i = \left\lvert x_i \right \rvert + \left\lvert x_i \right \rvert$, $a_1=\frac{\ln(2)}{5}$, and $a_2=\frac{\ln(2)}{0.5}$.
\smallskip

\subsubsection{Model Architecture}\label{modelarch} Our DRL controller should be sufficiently small to keep verification time reasonable and sufficiently large to be able to learn the necessary behavior. 
We found that reducing the hidden layer widths from 256 neurons to 20 neurons, while maintaining two hidden layers, acheives a good balance between verification time and expressive power.  Also, we swap the tanh activation functions for ReLU activation functions since ReLU is supported by most neural network verification tools (such as Marabou).
\section{Using $k$-induction For Liveness Guarantees}
\label{sec:kinduction}



In this section, we present an approach for scalably verifying a liveness property for the 2D docking problem presented in Section \ref{sec:dockingproblem} using $k$-induction. We describe the conceptual approach, the experimental framework, and the results.


\subsection{Proving Liveness by $k$-induction}
In order to apply $k$-induction, we must find a way to reduce a liveness property to a $k$-inductive property.  Typically, this is done by finding a \emph{ranking function}, a function with a well-founded co-domain, which can be shown to always be decreasing by $k$-induction.

For the spacecraft, an obvious choice for a ranking function is the distance from the deputy to the chief.  In order to make the function easier to reason about, we use a linear proxy function for the actual distance, namely the Manhattan distance.  Unfortunately, it is not the case that this measure always decreases, as the spacecraft may move away from the target.

Thus, we instead propose a property that ensures the spacecraft eventually starts moving towards the target.  The property
is expressed as a logical disjunction: after $k$ steps, either the Manhattan distance decreases or the magnitude of the velocity decreases.  Again, we approximate the velocity magnitude by the $L^1$ norm, the sum of the absolute values of $\dot{x}$ and $\dot{y}$.  Formally, if the current state is $(x_0,y_0,\dot{x}_0,\dot{y}_0)$ and the future state after $k$ steps is $(x',y',\dot{x}',\dot{y}')$, we must show:

\begin{equation}\label{eq:ind}
(|x'| + |y'|) – (|x_{0}| + |y_{0}|) < –\epsilon \quad \bigvee \quad 
(|\dot{x}'| + |\dot{y}'|) – (|\dot{x}_{0}| + |\dot{y}_{0}|) < –\epsilon,
\end{equation}
where $\epsilon$ is some positive value. 

\newtheorem{proposition}{Proposition}

\begin{proposition}
If property \eqref{eq:ind} holds (for some $k$) for every state, then eventually the spacecraft will be moving towards the goal (i.e., the $L^1$ norm of the position will decrease).
\end{proposition}

\paragraph{Proof.} 
Suppose that from some starting state, $(x_0,y_0,\dot{x}_0,\dot{y}_0)$, the spacecraft follows a trajectory that never moves towards the goal in the sense that the $L^1$ norm never decreases.  Let $(x_i,y_i,\dot{x}_i,\dot{y}_i)$ be the state after $i$ time steps.  This means that for all $i$, $|x_i|+|y_i| \le |x_{i+1}| + |y_{i+1}|$.  Let $V_i=|\dot{x}_i|+|\dot{y}_i|$.  By \eqref{eq:ind}, we know that for each $V_i$, there must be some $k$, such that $V_{i+k} - V_i < -\epsilon$.  Thus, for any $n$, we can construct a sequence $V_{j_0},V_{j_1},V_{j_2},\dots V_{j_n}$ such that $j_0=0$ and $V_{j_i} - V_{j_{i+1}} > \epsilon$.  If we then take $n > V_0/\epsilon$, we get that $V_{j_n} < 0$, which is impossible.
\qed

\paragraph{Algorithm.} We verify \eqref{eq:ind} using Algorithm \ref{alg:k_ind_alg_pseudo}.  We gradually increase $k$ until the property holds, a maximum of $k=k_{max}$ is reached, or a timeout is exceeded.



\begin{algorithm}
\caption{Algorithm for $k$-induction.}
\label{alg:k_ind_alg_pseudo}
\begin{algorithmic}[1]
\REQUIRE Bounds on state components $x_{0}, y_{0}, \dot{x}_{0},\dot{y}_{0}$, values for $k_{min}, k_{max}$
\ENSURE If result = UNSAT, then property \eqref{eq:ind} holds for all states within the defined bounds.
    \FOR{each $k \in [k_{min}, k_{max}]$}
        \STATE Verify the negation of the distilled property: 
            \ensuremath{\neg \left ( \begin{aligned}
                & (|x'| + |y'|) – (|x_{0}| + |y_{0}|) < –\epsilon \\ 
                & \bigvee \\ 
                & (|\dot{x}'| + |\dot{y}'|) – (|\dot{x}_{0}| + |\dot{y}_{0}|) < –\epsilon )
                \end{aligned} \right)}
            
            \IF{UNSAT}
                \STATE result = [UNSAT, $k$]
                \STATE \emph{break};
            \ELSE
                \STATE result = [SAT, $k$, counterexample $k$-step trajectory].
            \ENDIF
    \ENDFOR
\RETURN{result} 
\end{algorithmic}
\end{algorithm}


Input bounds for the state space can be chosen according to the problem specification.  It is also important to note that different $k_{min}$ and $k_{max}$ values can be chosen.
In practice, in order to make the verification more tractable, we  first split the state space into subregions, then call the algorithm on each subregion.
For each subregion of the state space, we explore values of $k$ from $k_{min}$ to $k_{max}$.  For each $k$, a neural network verifier is invoked to check if the negation of the property holds after $k$ steps.  There are three possible results of the algorithm.

\begin{enumerate}
\item If the negation of the property is satisfiable for each $k$, the algorithm returns \textit{SAT} along with a counter-example.
\item If the negation of the property is unsatisfiable for some $k$, this means that the property holds for that value of $k$.  In this case, the algorithm returns \textit{UNSAT} together with the value of $k$ for which unsatisfiability was determined.  In this case, verification of the region is complete. 
\item If a predefined timeout is exceeded, the algorithm terminates and a timeout result is returned.
\end{enumerate}



\textbf{Experimental Setup}. 
%
We use Marabou for the neural network verification step. 
We set the following parameters for Marabou: \textit{``verbosity=0, timeoutInSeconds=5000, numWorkers=10, tighteningStrategy=``sbt'', solveWithMILP=True''}.  Marabou also requires a back-end linear programming engine.  We use Gurobi 9.5.

We start with positional bounds of $|x|,|y|\in[-25,25]$ and velocity bounds of $\dot{x},\dot{y}\in[-0.2,0.2]$).  We initially divide these into 25 subregions by focusing on $5 \times 5$ regions in the positional space.  A subregion is further subdivided if Algorithm \ref{alg:k_ind_alg_pseudo} times out.  We set $k_{min}$ to 1, $k_{max}$ to 20, and use a timeout of 1.4 hours for each loop iteration (i.e., 30 hours if all values of $k$ time out).

\textbf{Results}.
We end up with 71 subregions.  For each subregion, Algorithm \ref{alg:k_ind_alg_pseudo} returns UNSAT.  The minimum returned value for $k$ is 1, the maximum is 12, the average is 5, and the median is 3.

Notably, regions close to the goal region are more difficult: they require more subregions and take longer, whereas regions more distant can sometimes be verified without utilizing additional subregions.  The minimum runtime (in seconds) for any subregion is 0.02, the maximum is 4295.86, the average is 193.62, and the median is 1.76.

As a sanity check, we validated our results experimentally by running a simulation framework. Starting from randomly sampled points in the state space, we confirmed that the $k$-inductive property holds on the trajectory starting at each point.  These checks also succeeded.

\paragraph{Discussion.}
Initially, we applied our approach to the neural network controller described in~\cite{ravaioli2022safe}.  The original network topology (two hidden layers with 256 nodes each) resulted in lengthy verification times.  Moreover, for many regions, the verification failed: we discovered counter-examples for all tested values of $k$.


\begin{figure}
    \centering
\begin{subfigure}[b]{0.30\textwidth}
\includegraphics[width=\textwidth]
{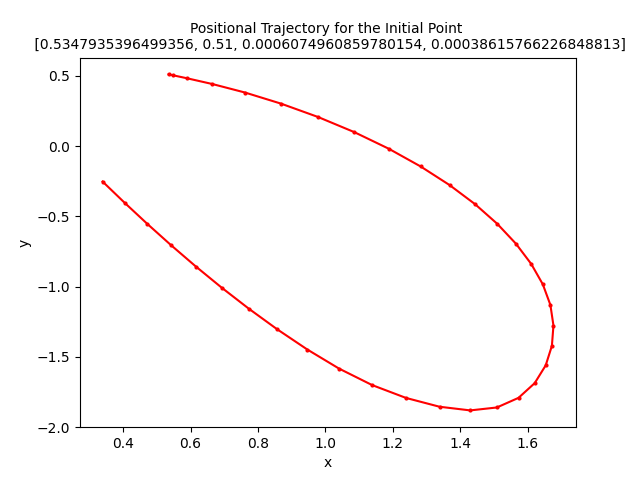}
\caption{Initial neural network.}
\label{fig:image_orig_NN}
\end{subfigure}
\begin{subfigure}[b]{0.30\textwidth} \includegraphics[width=\textwidth]{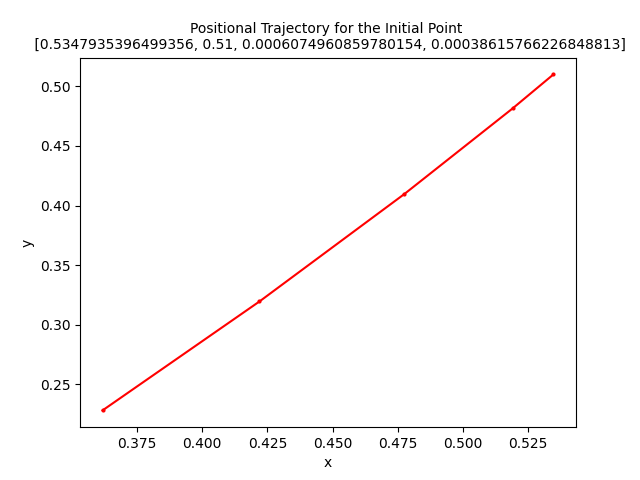}    
\caption{Retrained neural network.}  
\label{fig:image_retrained_NN}
\end{subfigure} \caption{Design for Verification: An initial controller trajectory compared to a final controller trajectory, with the same initial state. The final controller has a more direct trajectory which is more conducive to verification via $k$-induction.}
\label{fig:orig_vs_retrained}
\end{figure}

 Figure~\ref{fig:image_orig_NN} shows an example counterexample trajectory from the original neural network. The starting state is $[x = 0.5347935396499356, y=0.51, \dot{x}=0.00038615766226848813, \dot{y}=0.00038615766226848813]$. The controller moves steadily away from the goal, and only after many steps turns the spacecraft around to move towards the goal.

 Such trajectories provided motivation for the design changes mentioned in Section~\ref{subsec:DNN_setup}.  In particular, the changes to the reward function strongly incentivize the controller to move towards the goal region.  Figure \ref{fig:image_retrained_NN} shows the trajectory using the verified controller, starting from the same starting state.  Note how the spacecraft moves nearly directly towards the goal region.

The successful verification of \eqref{eq:ind} is not sufficient to establish that the deputy eventually reaches the chief.  We would need to establish a second property, namely that once the spacecraft is moving towards its goal, it always  gets closer (by at least some $\epsilon$) within $k$ steps.  Let $x_i,y_i$ be the position $i$ steps from some starting position $(x_0,y_0)$.  This can be formalized with the property:

\begin{align}
\begin{split}
(|x_1|+|y_1|)–(|x_0|+|y_0|) < 0 \implies\\
\quad \quad \exists\,k.\: (|x_k|+|y_k|)–(|x_0|+|y_0|) < –\epsilon.
\end{split}
\end{align}

\noindent
Formally verifying this property is left to future work.
 

%
%

\subsection{An Alternative Approach using Polar Coordinates}
Before moving to the Manhattan distance, we explored an alternative approach using polar coordinates, which allows the $L^2$ norm to be used directly in the invariant while maintaining linearity.  More specifically, if $r$ is the distance to the origin and $\theta$ is the angle from the $x$-axis, then we can write the equivalent of property \eqref{eq:ind} as:

\begin{equation}\label{eq:polar}
r' - r < -\epsilon \vee \dot{r}' - \dot{r} < -\epsilon.
\end{equation}

Note how much simpler property \ref{eq:polar} is compared with property \eqref{eq:ind}.  However, there remain two challenges: training a polar controller and converting the dynamics to polar coordinates.

Training a controller for the polar system is not straightforward; it requires complex parameter changes, for example, adjusting the learning rate, observation vector order, and the length and normalization constants.  However, these challenges are ultimately solvable, and we were able to train a network that takes polar coordinate inputs.  The output is still $F_x$ and $F_y$, as we did not envision changing the physical spacecraft system.

The second challenge proved more difficult.  We needed a way to calculuate new values of $r$ and $\theta$, given current values of $r$, $\theta$, $\dot{r}$, and $\dot{\theta}$, as well as $F_x$ and $F_y$.
We did not find closed-form solutions in the literature for the Clohessy–Wiltshire Equations utilizing polar coordinates.  We thus converted equations~\eqref{eq:cw1} through ~\eqref{eq:cw4} to polar coordinates using the standard conversion equations:

\begin{align}
    x = r \cos \theta, \quad 
    y = r \sin \theta, \quad
    r = \sqrt{x^2 + y^2}, \quad 
    \theta = \tan ^{-1} \frac{y}{x}
\end{align}

We encoded the derivation of the equations directly in \textit{Python}, which allowed us to confirm in simulation that our polar neural network had behavior  similar to that of the original model.  However, attempting formal verification with the new dynamics proved difficult.  The new dynamics are highly non-linear.
We attempted to use the \textit{OVERT} tool\footnote{\href{https://github.com/sisl/OVERT.jl}{https://github.com/sisl/OVERT.jl}} for the purpose of linearizing $r$ and $\theta$. 
However, the results were too complex and ultimately unsuccessful.  It was at this point that we decided to instead use the $L^1$ norm and revert to standard rectangular coordinates.

We report this effort here in order to highlight both the potential benefits and pitfalls of using a different coordinate representation.  If the dynamics had been more tractable in polar space, this would have been an attractive direction.


\section{Alternate Verification Approaches} 
\label{sec:certificate}
While exploring the $k$-induction approaches described above, we concurrently explored an alternative approach using Neural Lyapunov Barrier certificates.  The results of that effort represent the most complete verification results we have obtained to date and are reported in~\cite{mandal2024formally}. Here, for convenience, we review that approach at a high level and present some details not reported there.  We also discuss several reachability-based approaches, which we also applied to the 2D docking problem, but which were, ultimately, unsuccessful.

\subsection{RWA Certificates} \label{cert:RWA}

\begin{definition}\label{def:rwacert}
A function $\certfun:\states\mapsto \mathbb{R}$ is an RWA certificate for the task defined in Definition~\ref{def:reachavoid} if there exist some $\alpha > \beta \ge \gamma$ and $\epsilon > 0$, such that the following constraints are satisfied. %

\begin{align}
& \forall\,\state\in\states. && \certfun(\state) \ge \gamma
\label{eq:CLBFcond0}\\
& \forall\,\state\in\states_I. && \certfun(\state) \le \beta
\label{eq:CLBFcond1} \\
& \forall\,\state\in\states\setminus\states_G. && \certfun(\state) \le \beta \rightarrow \certfun(\state) - \certfun(\trans(\state,\ctrlfun(\state))) \geq \epsilon 
\label{eq:CLBFcond2} \\
& \forall\,\state\in\states_U. && \certfun(\state) \ge \alpha \label{eq:CLBFcond3}
\end{align}
\end{definition}
Any tuple of values $(\alpha,\beta,\epsilon,\gamma)$ for which these conditions hold is called a \emph{witness for} the certificate.\footnote{These constraints are similar to those in ~\cite{EdwPerAba2023} but are specific to discrete-time systems and do not place constraints on a compact safe set, opting to use an unsafe set instead. }
RWA certificates provide the following guarantee.
\begin{lemma}
If \certfun is an RWA certificate for a dynamical system with witness $(\alpha,\beta,\epsilon, \gamma)$, then for every trajectory $\traj$ starting from a state $\state\in\states\setminus\states_G$ such that $\certfun(\state)\le \beta$, $\traj$ will eventually contain a state in $\states_G$ without ever passing through a state in $\states_U$.
\end{lemma}

\noindent
We use reinforcement learning to jointly train neural networks for both the controller and the corresponding RWA certificate.

\paragraph{RWA Training Loss.}
The training objective for RWA certificates is described below:
\begin{align}
O_s &= c_s \sum_{i \,|\, \state_i \in \states_I} \frac{\text{ReLU}(\delta_1 + \certfun(\state_i) - \beta)}{\sum_{i \,|\, \state_i \in \states_I} 1} \label{eq:ReluOne:1} \\
O_d &= c_d \!\!\!\!\!\!\sum_{i \,|\, \state_i \in \states \setminus (\states_U \cup \states_G), \certfun(\state_i) < \beta} \frac{\text{ReLU}(\delta_2 + \epsilon + \certfun(\state'_i) - \certfun(\state_i))}{\sum_{i \,|\, \state_i \in \states \setminus (\states_U \cup \states_G), V(\state_i) < \beta} 1} \label{eq:ReluTwo:2}\\
O_u &= c_u \sum_{i \,|\, \state_i \in \states_U} \frac{\text{ReLU}(\delta_3 - \certfun(\state_i) + \alpha)}{\sum_{i \,|\, \state_i \in \states_U} 1} \label{eq:ReluFive:1} \\
O &= O_s + O_d + O_u \label{eq:ReluFour:4}
\end{align}

Equation~\eqref{eq:ReluOne:1} penalizes deviations from constraint~\eqref{eq:CLBFcond1}, Equation~\eqref{eq:ReluTwo:2} penalizes deviations from constraint~\eqref{eq:CLBFcond2}, and Equation~\eqref{eq:ReluFive:1} penalizes deviations from constraint~\eqref{eq:CLBFcond3}. We incorporate parameters $\delta_1 > 0$, $\delta_2 >0$, and $\delta_3 > 0$, which can be used to tune how strongly the certificate over-approximates adherence to each constraint.
Similarly, constants $c_s$, $c_d$, $c_u$ can be used to tune the relative weight of the two objectives. The final training objective $O$ in \eqref{eq:ReluFour:4} is what the optimizer seeks to minimize, by using stochastic gradient descent (SGD) or other optimization techniques. 

\paragraph{$\gamma$ lower bound.}
It is important to note that the RWA training objective does not explicitly penalize deviations from Equation~\eqref{eq:CLBFcond0}. Instead, because $\certfun$ is implemented as a neural network using floating-point arithmetic, it has only a finite number of possible inputs and outputs, so Equation~\eqref{eq:CLBFcond0} must hold for some $\gamma$. In practice, we can use Marabou to find $\gamma$ by doing a linear search for the minimum value of $\certfun$: we simply set $\gamma$ to some initial value, say $\alpha$, then repeatedly check $\exists\,\state.\: \certfun(\state) < \gamma$, updating $\gamma$ with the new value each time the query is satisfiable, and repeat until the query is unsastisfiable.

\paragraph{Sampling from $\states_U$ and $\states \setminus \states_G$.}
\label{subsec:sampling}
While $\states_I$ is typically defined as having both upper and lower bounds on state variables, this is not the case for $\states_U$, which often has only lower bounds on state variables (this is the case, for example, for the 2D docking problem defined in Section \ref{sec:dockingproblem}).

However, during training, we do impose an upper bound on the states sampled from $\states_U$. Specifically, if the controller operates over $n$-dimensional states $\state = [x_1, x_2,.., x_n]$, we sample points satisfying the following constraints:
\begin{align}
    (x_1 > p_1) \vee (x_2 > p_2) \vee ... \vee (x_n > p_n)\label{eq:sample:1}\\
    (x_1 < p_1+\gamma_1) \wedge (x_2 < p_2 + \gamma_2) \wedge ... \wedge (x_n < p_n + \gamma_n) \label{eq:sample:2}
\end{align}
Here, \ref{eq:sample:1} represents the (given) lower bounds on the unsafe region $\states_U$, and $\gamma_1,...,\gamma_n$ are chosen to be strictly greater than 0.


A similar issue arises when sampling from $\states \setminus \states_G$.  This can often be solved simply by sampling instead from $\states \setminus (\states_G \cup \states_U)$, as the lower bounds on variables in $\states_U$ then create upper bounds for the sampling step.


\paragraph{Masking out $\states_U$.}
For objective \ref{eq:ReluTwo:2}, if $\state'_i$ lies in $\X_U$, we replace the actual value of $V(x'_i)$ with $\alpha$. This is because we learn correct functional behaviors of $\X_U$ through objective \ref{eq:ReluFive:1} regardless, and thus using the actual value of $V(x'_i)$ would lead to unnecessary training effort and excessive penalties.

\paragraph{Certificate Warmup.}
To improve training, the objective is used to train the certificate $V$ alone for a few iterations, after which training includes both the certificate and the controller. This is done to avoid erratic training of the controller when $V$ has random weights.

\paragraph{RWA Verification.}\label{subsec:verif}
In order to obtain formal guarantees, we use Marabou to formally verify the constraints in~\Cref{def:rwacert}.
Verification of RWA constraints is generally straightforward, but we have to similarly bound $\states_U$ and $\states \setminus \states_G$ to verify constraints \ref{eq:CLBFcond3} and \ref{eq:CLBFcond2} respectively. Instead of using $\states \setminus \states_G$ as the input space for \ref{eq:CLBFcond2}, we use instead $\states \setminus (\states_G \cup \states_U)$, which provides the same guarantees.
Moreover, instead of using $\states_U$ as the input space for \ref{eq:CLBFcond3}, we use the bounded space, call it $\states^S_U$, used for data sampling.  To ensure this provides the same guarantees, we check that no states beyond the upper bound of $\states^S_U$ are reachable.

Instead of encoding verification as a single property passed to the DNN verifier, verification is partitioned into muliple queries. This is done by paritioning the input space in the original property into equally sized smaller state spaces, over which the same property is checked. This helps avoid unreasonably long verification times that can occur with a large monolithic query.


\paragraph{Retraining.}
If any of the RWA verification checks return counterexamples, these are used to augment the training data set, and then training is done again.  This process repeats until no more counterexamples are found.  We weight counterexamples more heavily in the objective function \ref{eq:ReluFour:4} (compared to points in the initial training dataset) in order to focus the training on removing the counterexamples.

\paragraph{Results and Analysis.}
As shown in prior work in \cite{mandal2024formally}, RWA certificates can provide liveness and safety guarantees for the 2D spacecraft docking problem defined in Section \ref{sec:dockingproblem}.  More details and a pointer to the code can be found in \cite{mandal2024formally}.

\subsection{Reachability Analysis Approaches}
\label{reachability}
In this subsection, we discuss approaches based on \emph{reachability analysis}. While these approaches were ultimately unsuccessful on the case study problem outlined in section \ref{sec:dockingproblem}, we still mention them here, as the reasons for their failure may be of interest, and they may be useful on other problems.

\paragraph{Forward-tube and Backward-tube Reachability.}\label{3:4}
Forward-tube and backward-tube reachability attempt to generate a path over abstract state spaces (i.e., sets of states) from the starting state space to the goal state space. At each step along the abstract path, we check that every state in the abstract state set meets any safety guarantees.

In forward-tube reachability, a starting set of states $\states^0_F$ and step size $k$ is defined. Then, a set of states $\states^1_F$ is constructed such that all states reachable from $\states^0_F$ in $k$ steps are contained within $\states^1_F$. This process is continued, and additional sets of states $\states^{i+1}_F$ are constructed, each with the property that they contain the states reachable from $\states^{i}_F$ in $k$ steps.  If at some point, the constructed set is a subset of the goal region, then the liveness property is ensured. However, it can be very challenging to find a sequence of sets of states $\states^{i}_F$ that eventually lead to a subset of the goal region.  This was the case for the spacecraft example.

On the other hand, in backward-tube reachability, we start with $\states^0_B$ set equal to the goal states and define a step size $k$. Then, a set of states $\states^1_B$ is constructed such that all states reachable from $\states^1_B$ in $k$ steps are contained within $\states^0_B$.
Again, this process can be repeated until the set of states includes the initial states.  A difficulty with this approach is computing a sufficiently large previous set of states at each step.

\begin{figure}[t]
    \centering
    \includegraphics[width=0.4\textwidth]{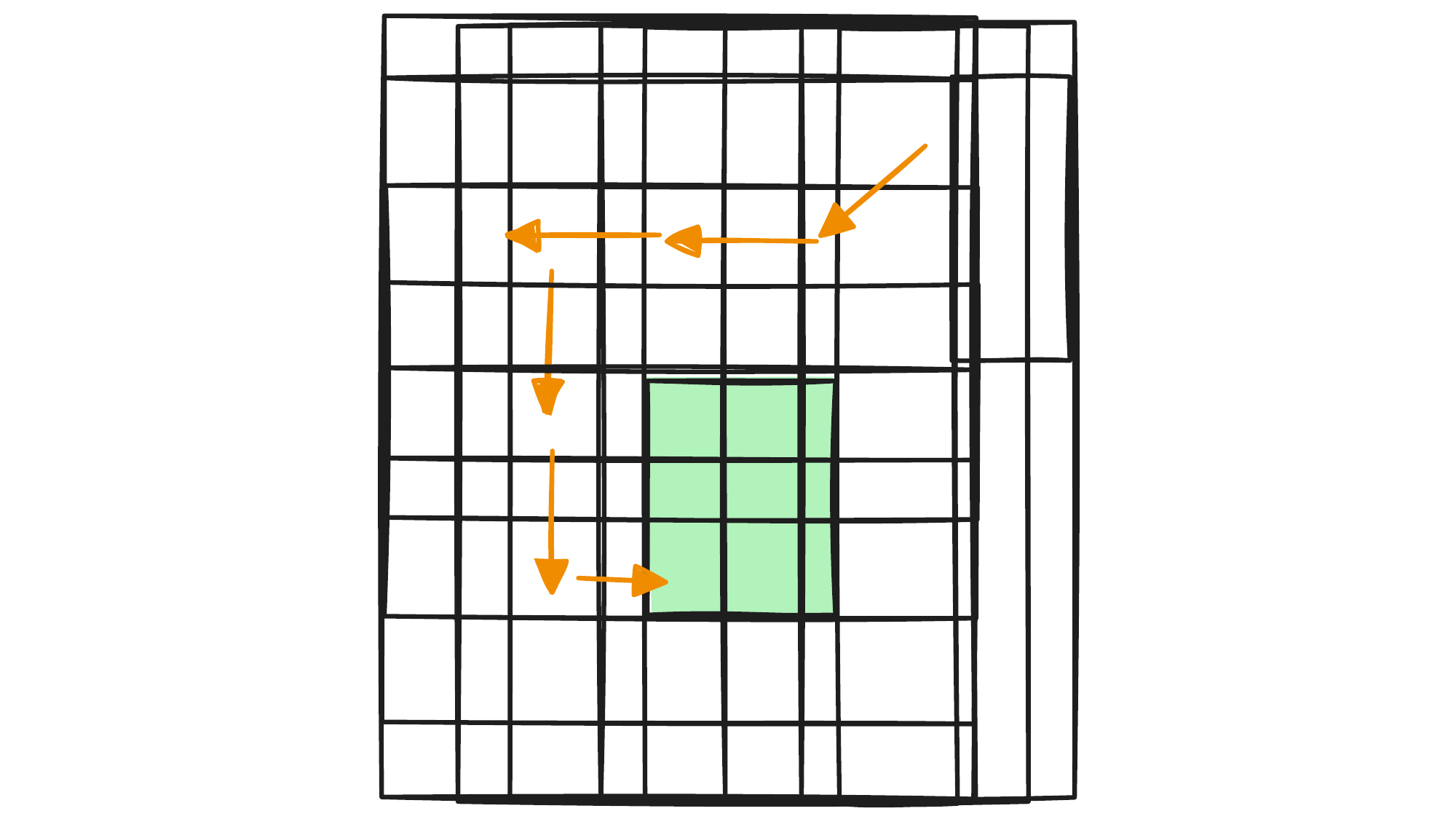}
    \caption{Grid reachability, with a cell navigating towards the docking region (in green)}
    \label{fig:gridFailedAttempt}
\end{figure}

\paragraph{Grid Reachability. }\label{3:3}
Grid reachability is a process which first partitions a bounded subset of the state space into cells, then computes a directed graph where each cell is a vertex, and each directed edge $(a,b)$ denotes that vertex $b$ is reachable from vertex $a$ in $k$ steps, for a specific $k$, as shown in Fig. \ref{fig:gridFailedAttempt}. The goal is to show that for all paths constructed from cells in the defined initial state space, a goal region reachable.  However, to ensure liveness, it is also necessary to show that the graph has no cycles and that it is not possible to reach any cells beyond the partitioned state space.


We applied this technique to the spacecraft example.
A challenge is preventing self-cycles in the graph.  One strategy for doing this is to construct cells where at least one velocity component never changes sign.  It is easy to see that for such cells, the spacecraft cannot remain in the cell forever, so we can ignore self-loops on such cells.  For cells containing a velocity sign-change, we use a very narrow velocity range, narrow enough to ensure that the spacecraft leaves the range in $k$ steps.  It is also desirable to limit the number of cells reachable from a given cell, to avoid the need to do many reachability checks.  This can be ensured by making the cells large enough that it is impossible to cross more than one cell in a single set of $k$ steps.


\newcommand{\IS}{\ensuremath\mathit{IS}\xspace}

\begin{algorithm}[t]
\caption{\textsc{Applying Grid Reachability}\label{alg:gridreach}
}
Let $\IS$ be the input space\\
Let $k$ be the step size\\
Divide $\IS$ into cells $ C = c_0,c_1,...,c_n$\\
Let vertices $V = C$\\
Initialize edge set $E$ to be the empty set\\
$i = 0$\\
\For {$i \le n$}{
Denote set of adjacent cells to $c_i$ as $C_r$\\
Add $c_i$ to $C_r$ if self-cycles are possible\\
\For {$c_r \in C_r$}{
\If {$c_r$ is reachable from $c_i$ in $k$ steps}{
Add directed edge $(c_i,c_r)$ to $E$\\
}
}
$i = i+1$
}
Let $G := (V,E)$\\
Check for cycles in $G$\\
\If {$G$ is acyclic}{
Determine cells $C_s$ with no paths leaving input space\\
\Return $C_s$ as cells meeting liveness property\\
}
\end{algorithm}


\paragraph{Analysis of Grid Reachability.}
We applied grid reachability to a state space with $x,y\in [-10,10]$  and $\dot{x},\dot{y}\in [-1.6,1.6]$ using Algorithm \ref{alg:gridreach}. A binary search was conducted using Marabou to determine cell bounds such that cells could only reach adjacent cells. The step size $k$ was chosen to be $1$. 

\begin{figure}[t]
  \centering
\includegraphics[width=1\columnwidth]{
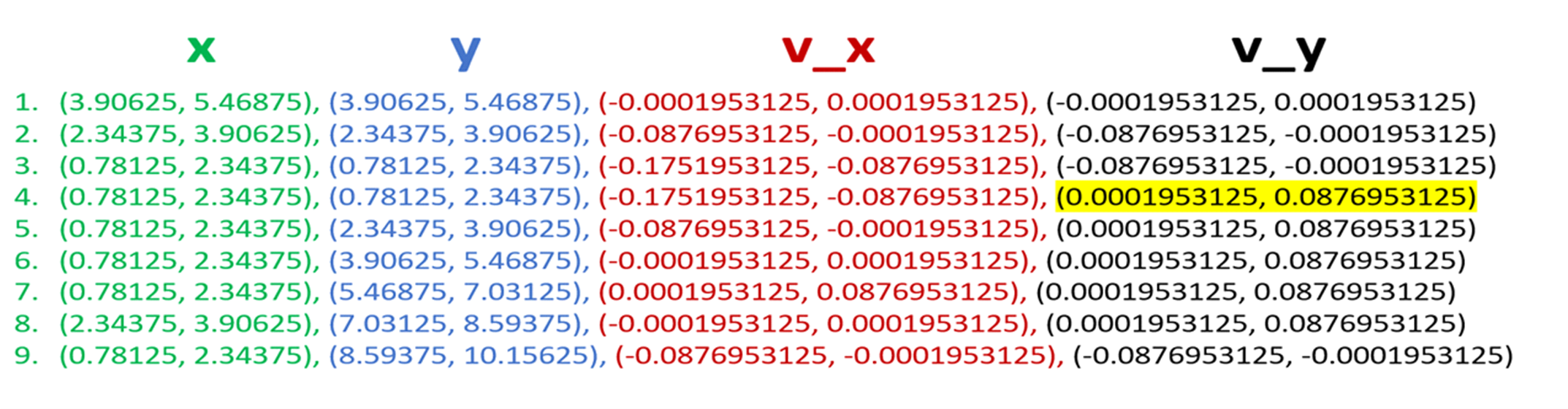}
  \caption{Spurious trajectory with grid reachability}
  \label{fig:gridreachfail}
\end{figure}

We found a variety of cycles of increasing lengths, even as cells were divided further in an attempt to refine the grid abstraction. Moreover, we found that all cells had paths leaving the input space.  We showcase one such trajectory of cells with this behavior in Fig. \ref{fig:gridreachfail}. In this trajectory, we see that for the first three steps, the velocity component ranges are negative, thereby guiding the spacecraft towards the goal region, but there is a path from cell 3 to cell 4 that induces a positive velocity component, allowing the path to diverge.

Ultimately, the grid abstraction does not lend itself well to the liveness task because such spurious paths are difficult to rule out.  While further refinement of the grid approach is possible and could eventually yield a workable approach, we determined that the complexity and difficulty were too high, and abandoned it in favor of the certificate approach mentioned earlier.

\section{Conclusion}
\label{sec:Conclusion}

We have presented methods for verifying safety and liveness properties for DRL systems using $k$-induction, Neural Lyapunov Barrier Certificates, and reachability analysis. We explore their effectiveness on a 2D spacecraft docking problem posed in previous work. For this problem, we show how a $k$-induction based approach can be used alongside a design-for-verification training scheme to provide liveness guarantees.  We also discuss how Neural Lyapunov Barrier Certificates can be used to provide both liveness and safety guarantees. While  reachability analysis ultimately did not provide any formal guarantees, we discuss the approach and its limitations. In future work, we plan to explore scaling these methods to more complex and realistic control systems.
\section{Acknowledgements}
\label{sec:Acknowledgements}
This work was supported by AFOSR (FA9550-22-1-0227), the Stanford CURIS program, the NSF-BSF program (NSF: 1814369, BSF: 2017662), and the Stanford Center for AI Safety.
The work of Amir was further supported by a scholarship from the Clore Israel Foundation.
We thank Thomas Henzinger (ISTA), 
Chuchu Fan (MIT), and Songyuan Zhang (MIT) for useful conversations and advice, which contributed to the success of this project.

{
\bibliographystyle{abbrv}
\bibliography{references}
}


\end{document}